%% file: main.tex
\documentclass[10pt,journal,compsoc]{IEEEtran}
\usepackage{amsmath,amsfonts}
\usepackage{algorithmic}
\usepackage{algorithm}
\usepackage{array}
\usepackage[caption=false,font=normalsize,labelfont=sf,textfont=sf]{subfig}
\usepackage{textcomp}
\usepackage{stfloats}

\usepackage{booktabs} 

\usepackage{url}
\usepackage{verbatim}
\usepackage{graphicx}
\usepackage{mwe}    
\usepackage{cite}
\usepackage[table,xcdraw]{xcolor}
\usepackage{makecell}
\usepackage{multirow}
\usepackage{hyperref}
\hypersetup{
    colorlinks=true,    
    linkcolor=blue,     
    citecolor=blue,    
    filecolor=blue,  
    urlcolor=cyan,      
    pdfborder={0 0 0}   
}
\newcommand{\ssh}[1]{{#1}}
\newcommand{\clh}[1]{{#1}}
\newcommand{\hb}[1]{{#1}}

\begin{document}

\title{VistaGEN: Consistent Driving Video Generation with Fine-Grained Control Using Multiview Visual-Language Reasoning}

\author{Li-Heng Chen$^*$, Ke Cheng$^*$$^\dag$, Yahui Liu, Lei Shi, Shi-Sheng Huang, Hongbo Fu 
\thanks{$^*$ Equal Contribution.$^\dag$ Project Lead. Li-Heng Chen and Hongbo Fu are with Hong Kong University of Science and Technology. Ke Cheng, Yahui Liu, Lei Shi are with Meituan, Inc. Shi-Sheng Huang is with Beijing Normal University. Shi-Sheng Huang and Hongbo Fu are the corresponding authors. }
}



\IEEEtitleabstractindextext{
\begin{abstract}
Driving video generation has achieved much progress in controllability, video resolution, and length, but \hb{fails to} 
\ssh{support} fine-grained object-level controllability for \emph{diverse} driving videos, while preserving the spatiotemporal consistency, especially in long video generation. In this paper, we present a new driving video generation \hb{technique}, 
called VistaGEN, which 
enables fine-grained control of specific entities, including 3D objects, images, and text descriptions, while maintaining 
spatiotemporal consistency \clh{in} long video sequences. Our key innovation is the incorporation of multiview visual-language reasoning into the long driving video generation. \ssh{To this end,} we 
inject visual-language features into a multiview video generator to enable \emph{fine-grained} controllability. More importantly, we propose a multiview vision-language evaluator (MV-VLM)  
to intelligently and automatically evaluate spatiotemporal {consistency} of the generated content, \ssh{thus formulating a novel \emph{generation-evaluation-regeneration} closed-loop generation mechanism}. This 
mechanism ensures high-quality, coherent outputs, facilitating the creation of complex and reliable driving scenarios. Besides, \ssh{within the closed-up loop generation,} we introduce an object-level refinement module 
to refine the unsatisfied results 
evaluated from the MV-VLM and then \clh{feed them} back to the video generator for regeneration. Extensive evaluation shows that our VistaGEN achieves diverse driving video generation results with fine-grained 
controllability, especially for long-tail objects, and much better spatiotemporal consistency than previous approaches. 
\end{abstract}

}




\maketitle

\input{sec/0-intro}

\input{sec/1-relate_work}

\input{sec/2-method}
\input{sec/3-exp}

\input{sec/4-conclusion}

\bibliographystyle{IEEEtran}

\bibliography{main}

\end{document}

%% file: sec/0-intro.tex
\section{Introduction}

\IEEEPARstart{V}ideo generation for driving scenes~\cite{blattmann2023align,chen2025gs,huang2025survey} or streetviews~\cite{wang2025cinemaster,kim2025videofrom3d,deng2024streetscapes} serves as an essential technique for data collection and annotation, and can boost up many applications in autonomous driving~\cite{wang2024driving,yang2025instadrive,ren2025gen3c,yang2025x}, virtual city roaming~\cite{takeuchi2021gibson,huang2022real,tang2025aerial,yao2025magiccity}. It has drawn extensive research attention from computer graphics and vision communities, especially inspired by the recent success of diffusion models~\cite{croitoru2023diffusion,wang2025diffusion} or world models~\cite{agarwal2025cosmos,awais2025foundation}. 

Early approaches~\cite{hu2022model,gao2024enhance} 
incorporated Bird’s Eye
View (BEV) cues as semantic priors~\cite{swerdlow2024street} for 3D city modeling, and control the video generation using 2D geometric guidance~\cite{yang2023bevcontrol}. The rise of diffusion models expressively improved the video generation quality, especially exemplified by notable contributions such as DriveDreamer~\cite{wang2024drivedreamer}, Drive-WM~\cite{wang2024driving}, Magic-Drive~\cite{gao2023magicdrive}, and Panacea~\cite{wen2024panacea}. Recent works such as MagicDriver-2~\cite{gao2025magicdrive} further improved the video resolution and length by incorporating DiT~\cite{peebles2023scalable} and 3DVAE~\cite{yang2024cogvideox}. However, most of the previous driving video generation approaches highly rely on structure prompts (such as BEV, 3D boxes, HDMaps, and optical flow), without an effective ability for \emph{fine-grained} controllability of object-level manipulation. Although some recent attempts like DriveDreamer-2~\cite{zhao2025drivedreamer} 
combine large language models (LLMs)~\cite{hu2023gaia} for more diverse driving video generation, maintaining the spatiotemporal consistency for multiview video generation still remains challenging, especially for long driving video generation. 


In this paper, we propose a new driving video generation approach, called VistaGEN, which enables diverse driving video generation with fine-grained object-level controllability, while preserving much better spatiotemporal consistency of generated long video sequences. Our approach is inspired by the high resolution and long generation ability model~\cite{gao2025magicdrive}, and 
introduces effective multiview visual-language reasoning into the video diffusion process for much better fine-grained control ability and consistency than previous approaches~\cite{gao2025magicdrive,zhao2025drivedreamer,yang2025instadrive}. 

\begin{figure*}
    \centering
    \includegraphics[width=0.98\linewidth]{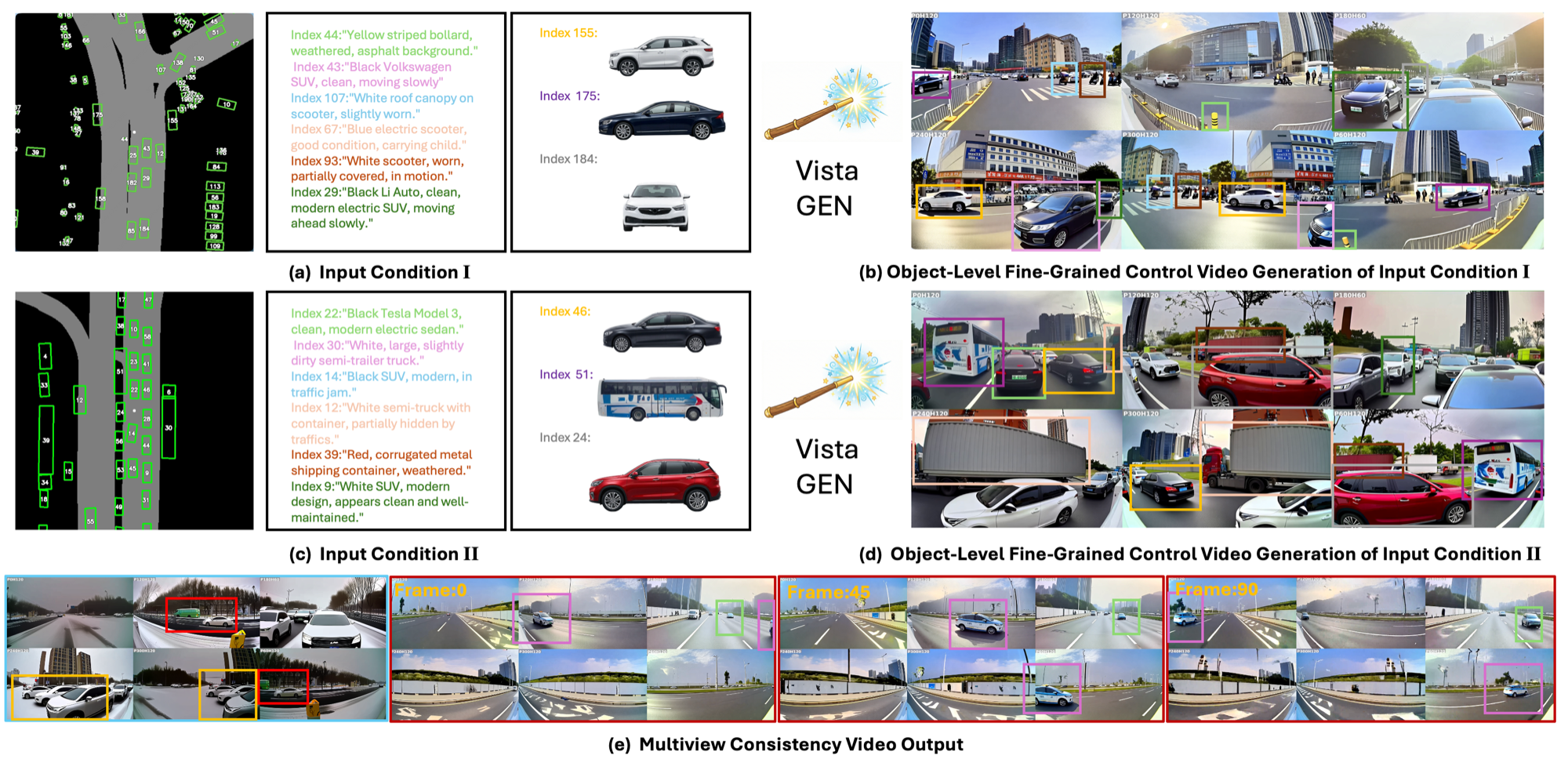}
    \caption{
    This paper introduces a new driving video generation technique, VistaGEN, which enables fine-grained control, like specifying 
    multiple objects' visual-language conditions (\ssh{roadmaps, text descriptors and visual appearance in (a) and (c), where the index numbers match the corresponding roadmaps in the right)}, with high-quality and coherent video generation outputs ((b) and (d)). Moreover, our VistaGEN can achieve spatiotemporal consistent generation (e) with both multiview consistency (colored red and orange in the left) and long-range temporal consistency (colored pink \ssh{and cyan} across different frames \ssh{respectively}) in the long video sequences. \ssh{Please refer to our demo video for more details.} 
    }
    \label{fig:teaser}

\end{figure*}

Specifically, we first fuse the visual-language features as a fine-grained prompt into a multiview video generator, enabling better 
fine-grained object-level control. Moreover, we introduce a multiview vision-language evaluator (Multi-VLM) to automatically evaluate the precision of multiview generation \hb{at the object level} 
and propose a subsequent object-level refinement. This results in 
a novel \emph{generation-evaluation-regeneration} closed-loop mechanism, enabling the preservation of 
the content consistency during the long-range video sequences, as shown in Fig.~\ref{fig:teaser}. \ssh{Besides, we also build up an object-level refinement module, which uses explicit 3D geometric cues to improve the object-level spatio-temporal coherence within the closed-up loop generation.} 

To evaluate the effectiveness of our VistaGEN, we conduct extensive experiments on public~\cite{caesar2020nuscenes} \ssh{and self-collected real-world }datasets, by comparing with 
state-of-the-art driving video generation approaches such as MagicDrive-V2~\cite{gao2025magicdrive} and DriveDreamer-2~\cite{zhao2025drivedreamer}. From the experiments, our VistaGEN can generate diverse driving videos with better fine-grained control (using CLIP-T, CLIP-I scores~\cite{chen2023clip2scene} accuracy metrics and MV-VLM Evaluation), and 
better spatio-temporal consistency than those previous approaches, while maintaining similar video resolution, length, and quality (using FVD and FID accuracy metrics) to MagicDriver-2~\cite{gao2025magicdrive}. To our best knowledge, our VistaGEN achieves 
the state-of-the-art fine-grained driving video generation performance while preserving the spatio-temporal consistency for long video sequences. 

%% file: sec/1-relate_work.tex
\begin{figure*}[htbp]
    \centering
    \includegraphics[width=0.98\linewidth]{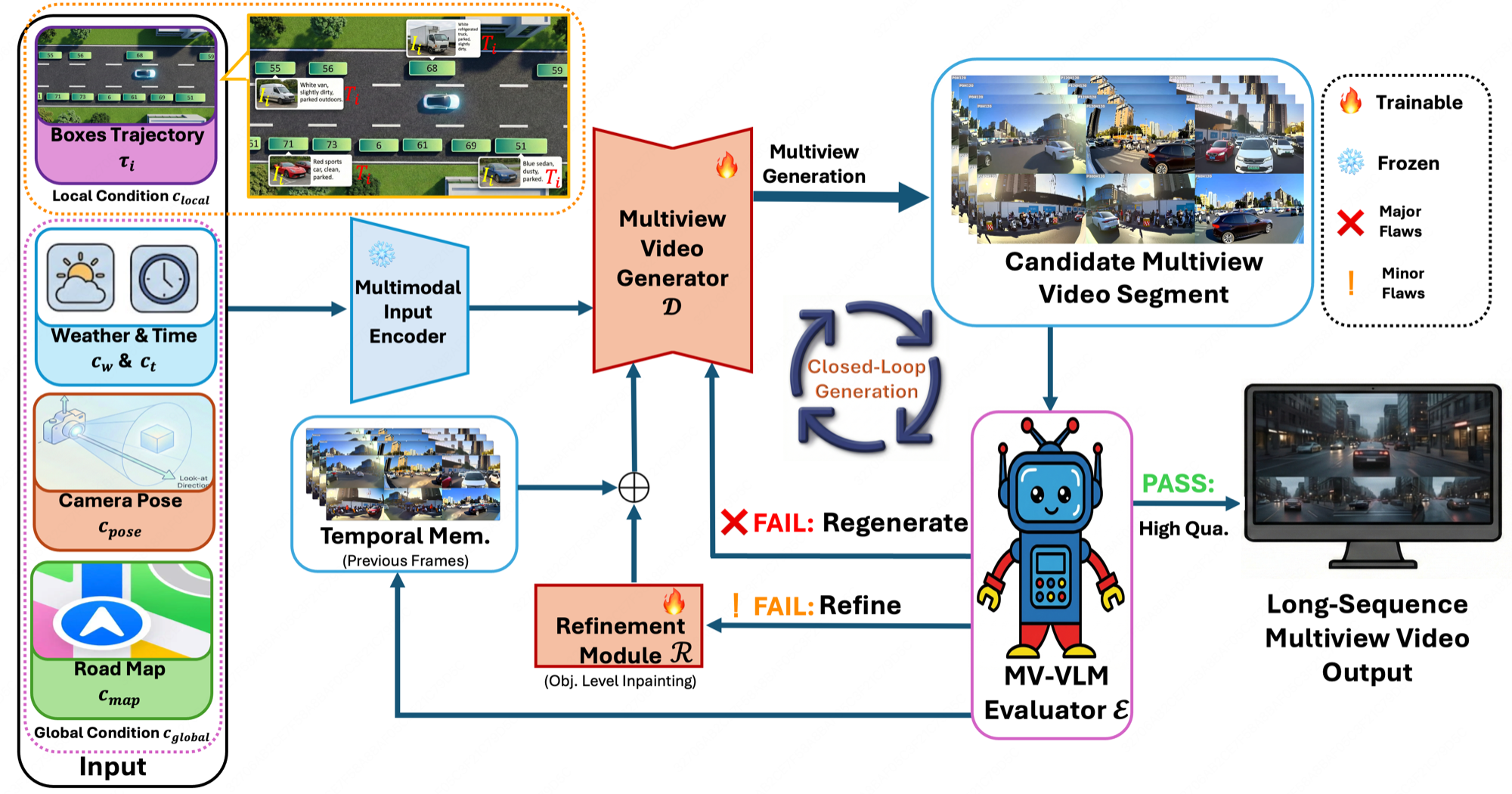}
    \caption{\textbf{The pipeline of VistaGEN.} Given frame descriptors as input, we hierarchically inject these signals as global and local scene control guidance via visual-language feature fusion for fine-grained controllability. Subsequently, we perform driving video generation via a multiview video generator $\mathcal{D}$. The generation is automatically evaluated by a multiview vision-language evaluator $\mathcal{E}$, followed by an object-level refinement module $\mathcal{R}$. This formulates a closed-loop mechanism to maintain spatiotemporal consistency for long video sequence generation. }  
    \label{fig:pipeline}
\end{figure*}

\section{Related Work}

Video generation is a classic problem in computer graphics~\cite{ma2025controllable,xing2024make,qu2025controllable} and is a pivotal technique for understanding the visual world. In this section, we will cover the most relevant
works below. 

\textbf{3D Reconstruction for Driving Scenes.} The recent success of neural implicit representation~\cite{park2019deepsdf,wang2021neus,huang2024neuralindicator, du2020sanihead} and neural rendering such as NeRF~\cite{mildenhall2021nerf} and 3DGS~\cite{kerbl20233d} has significantly improved the 3D reconstruction quality, and also enables large-scale scene reconstruction~\cite{lu2024scaffold,wang2024learning}. For driving scene reconstruction, S-NeRF++~\cite{10891659} introduces an enhanced NeRF 
for synthesizing large-scale scenes and moving vehicles. DrivingGaussian~\cite{zhou2024drivinggaussian} and its subsequent works~\cite{xiong2025drivinggaussian++} leverage dynamic 3D Gaussians to reconstruct complex moving objects in driving scenes. Recently, 
some works proposed to use explicit structural priors~\cite{wang2024drivedreamer,lu2025infinicube, zhang2020fast} 
for generative 4D driving scene reconstruction and synthesis. 
These 3D driving scene reconstruction approaches also motive subsequent works like large-scale 3D scene inpainting~\cite{cheng2024gaussianpro,chen2025gs}, novel view synthesis~\cite{meuleman2025fly} and attribution editing~\cite{du2025gs,li2025mpgs}. Our work is inspired by those previous 3D driving scene reconstruction approaches, but focuses on controllable driving scene generation in multiview video sequences with spatial and temporal consistency.

\textbf{Video Generation of Autonomous Driving.} Video generation for autonomous driving is an essential tool for data collection with various applications in training perception models~\cite{gao2023magicdrive}, testing~\cite{yang2025drivearena}, and scene reconstruction~\cite{gao2024magicdrive3d,zhao2025drivedreamer4d}. The pivotal criteria for driving video generation are realistic rendering quality and user-friendly controllability. 
The Diffusion models have improved controllable multi-view video generation with expressive driving scene rendering quality~\cite{wang2024driving,wen2024panacea}, while still lacking adequate resolution and frame count for policy testing~\cite{hu2023planning} and data engine applications~\cite{yang2025drivearena}. The recent work of MagicDriver-V2~\cite{gao2025magicdrive} expressively improved the video resolution and length of driving video generation with rich control prompts. But they still lack fine-grained controllability over video generation, preventing the generation of more diverse driving videos. \ssh{SubjectDrive~\cite{huang2025subjectdrive} explores the potential of generative models to produce vast quantities of freely-labeled data for autonomous driving.} DriveDreamer-2~\cite{zhao2025drivedreamer} proposes to inject large language models (LLM) into the video generation, thus enabling more user-friendly text-based video generation even at the instance level~\cite{yang2025instadrive}. However, temporal consistency, especially in long video sequences, could not be satisfactorily preserved by DriveDreamer-2, leading to non-coherent scene or object generation. In contrast to previous approaches, our VistaGEN pays special attention to the spatiotemporal consistency of long video generation while providing fine-grained object-level controllability.

\textbf{Driving-based Vision-Language Model.} Vision-Language models (VLMs)~\cite{radford2021learning,dong2025probing} have shown remarkable ability in advancing autonomous driving with multi-modal fusion to improve 3D scene perception~\cite{zhi2025lscenellm}, reasoning~\cite{chen2024spatialvlm}, and decision-making~\cite{zhai2024fine}. Unlike previous deep-learning-based perception models that rely on visual inputs, VLMs enable moving objects, such as vehicles, to describe their surroundings in natural language, thereby benefiting various applications in autonomous driving. For example, OpenScene~\cite{peng2023openscene} fuses 3D features with CLIP's image-text feature embeddings for better novel classes identification. Similarly, CLIP2Scene~\cite{chen2023clip2scene} combines the CLIP feature with a 3D segmentation model to achieve label-efficient 3D scene understanding. Besides, benefit from the strong reasoning abilities and world knowledge, 
notable works are proposed for high-level driving planning~\cite{mao2023gpt}, navigation~\cite{shao2024lmdrive}, and decision-making~\cite{wang2025omnidrive}. Our work is inspired by the expressive power of VLMs, and incorporates a multiview vision-language model based evaluator into the video generation process, thereby forming a closed-loop mechanism for fine-grained, consistent long-video generation.

%% file: sec/2-method.tex
\section{VistaGEN}

{Given a sequence of frame descriptions $\{\mathbf{C}\}$ for driving scenes, our VistaGEN aims to generate the corresponding driving video $\{I_{v,t}\} = \mathcal{G}(\mathbf{C} , \mathbf{z})$  from latent $\mathbf{z} \simeq \mathcal{N}(0,1)$, where $v \in \{0,...,V\}$ is the camera views. Here we also generate $V=6$ views similar to previous works~\cite{gao2025magicdrive} but in a closed-loop manner.} Specifically, our generation $\mathcal{G}$ contains three main modules, i.e., $\mathcal{G} = \{\mathcal{D} , \mathcal{E} , \mathcal{R}\}$, including a {multiview video generator $\mathcal{D}$}, multiview VLM evaluator $\mathcal{E}$, and an instance-level refinement module $\mathcal{R}$. Our key insight is to leverage the visual-language reasoning cues to $\mathcal{D}$ for fine-grained control and $\mathcal{E}$ for spatiotemporal consistency, respectively, and formulate a \emph{generation-evaluation-regeneration} loop into a unified end-to-end training framework for \emph{diverse} and \emph{consistent} driving video generation. The pipeline of our VistaGEN is shown in Fig.~\ref{fig:pipeline}. 

\subsection{Multi-View Video Generation}

To achieve {high resolution and long driving video generation,} we 
build a multi-view video generator $\mathcal{G}$ based on a {Multi-View Spatial-Temporal Diffusion Transformer (MV-STDiT)} within the compressed latent space of a 3DVAE inspired by recent work~\cite{gao2025magicdrive}. But for fine-grained controllability, we introduce a hierarchical condition injection 
mechanism {which fuses diverse control conditions with visual-language features injected}, enabling fine-grained and precise manipulation of the appearance and motion of dynamic agents {during video generation}.


\textbf{Hierarchical Condition Injection.} 
We decompose the control conditions $\mathcal{C}$ into \textit{Macro-level Global Scene Control} $\textit{c}_{global}$ and \textit{Micro-level Fine-grained Object Control} $\textit{c}_{local}$, injecting them into $\mathcal{G}$ via hierarchical cross-attention layers by fusing $\mathcal{C} = \mathbf{c}_{global} \cup \mathbf{c}_{local}$.

{For macro-level global scene control,} we combine the global environmental descriptions, including weather $c_{w}$, time of day $c_{t}$, and ego-vehicle pose $c_{pose}$, encoded as global embeddings to guide the overall rendering style and geometric perspective. Additionally, road map information $c_{map}$ is injected via a ControlNet-style \textit{Additive Branch} to ensure the structural accuracy of the road layout. Formally, we collect the $\mathbf{c}_{global} = \{ c_{w}, c_{t}, c_{pose}, c_{map} \}$ to model the macro-level global scene control.

{For fine-grained object-level representation,} 
unlike previous BEV layouts with only geometric positions lacking specific appearance descriptions (e.g., "a red Ferrari" or "a truck with specific livery"), we propose a multimodal object feature injection. Specifically, for the $i$-th object in the scene, with control signal is defined as $\mathbf{o}_i = (\tau_i, T_i, I_i)$ including:
(1) \textbf{Trajectory Layout ($\tau_i$),} the spatio-temporal trajectory of the object in BEV space, where we utilize Fourier Positional Encoding to calculate 3DBox-based geometric features $\mathbf{f}^{geo}_i$, (2) \textbf{Semantic Text Feature ($T_i$),} fine-grained textual descriptions (e.g., "A construction truck with a yellow mixer"), where we employ a pre-trained text encoder to extract semantic features $\mathbf{f}^{txt}_i$, and (3) \textbf{Visual Appearance Feature ($I_i$),} visual fidelity beyond textual description (e.g., maintaining consistent texture), where we 
utilize \textbf{SigLIP} \cite{tschannen2025siglip} 
to extract high-fidelity visual features $\mathbf{f}^{vis}_i = \text{SigLIP}(I_i)$ for each reference image $I_i$. 

\textbf{Visual-Language Feature Fusion.} 
A critical challenge in using 3D VAEs is the temporal compression, which can lead to misalignment between frame-wise control signals and compressed video latents. To address this, we use an {Object-Adapter} to fuse the multimodal features. {Specifically, we first fuse the visual-language features ($\mathbf{f}^{txt}_i, \mathbf{f}^{vis}_i$) into the structure-based geometric prompt ($\mathbf{f}^{geo}_i$) during the video diffusion process. This integration strategy significantly enhances the model's ability to achieve fine-grained object-level control via text prompts.} For each object, the final control embedding $\mathbf{e}_i$ is computed as:
\begin{equation}
\label{eq:e_i}
\mathbf{e}_i = \text{MLP}( \text{Concat}(\mathbf{f}^{geo}_i, \mathbf{f}^{txt}_i, \mathbf{f}^{vis}_i) ) + \mathbf{p}_{id},
\end{equation}
where $\mathbf{p}_{id}$ is a learnable identity positional encoding used to track the specific object instance across the long sequence. 
Consequently, we aggregate the control signals for all $M$ dynamic agents into the local condition set:
\begin{equation}
\mathbf{c}_{local} =\{ (\tau_i, T_i, I_i) \}_{i=1}^{M} = \{ \mathbf{e}_i \}_{i=1}^{M}.
\end{equation}
By injecting the unified conditions $\mathcal{C} = \mathbf{c}_{global} \cup \mathbf{c}_{local}$ into the multiview video generator, the final high-fidelity video $\hat{\mathbf{x}}$ is synthesized via the VAE decoder after the denoising process:
\begin{equation}
\hat{\mathbf{x}} = \mathcal{D}(\text{DiT}(\mathbf{z}_{init}, \mathcal{C}, t)).
\end{equation}

\subsection{Multiview VLM Evaluator}\label{mv-vlm evaluator}

One main problem for the multiview video generator is the consistency, for in scenes containing dense bounding box information or high-speed dynamic agents, it easily suffers from inconsistencies or visual artifacts. To mitigate this, we introduce an intelligent evaluator $\mathcal{E}$, instantiated by a Multiview Vision-Language Model (MV-VLM), to preserve the generation quality. 



\begin{figure}[t]
    \centering
    \includegraphics[width=\linewidth]{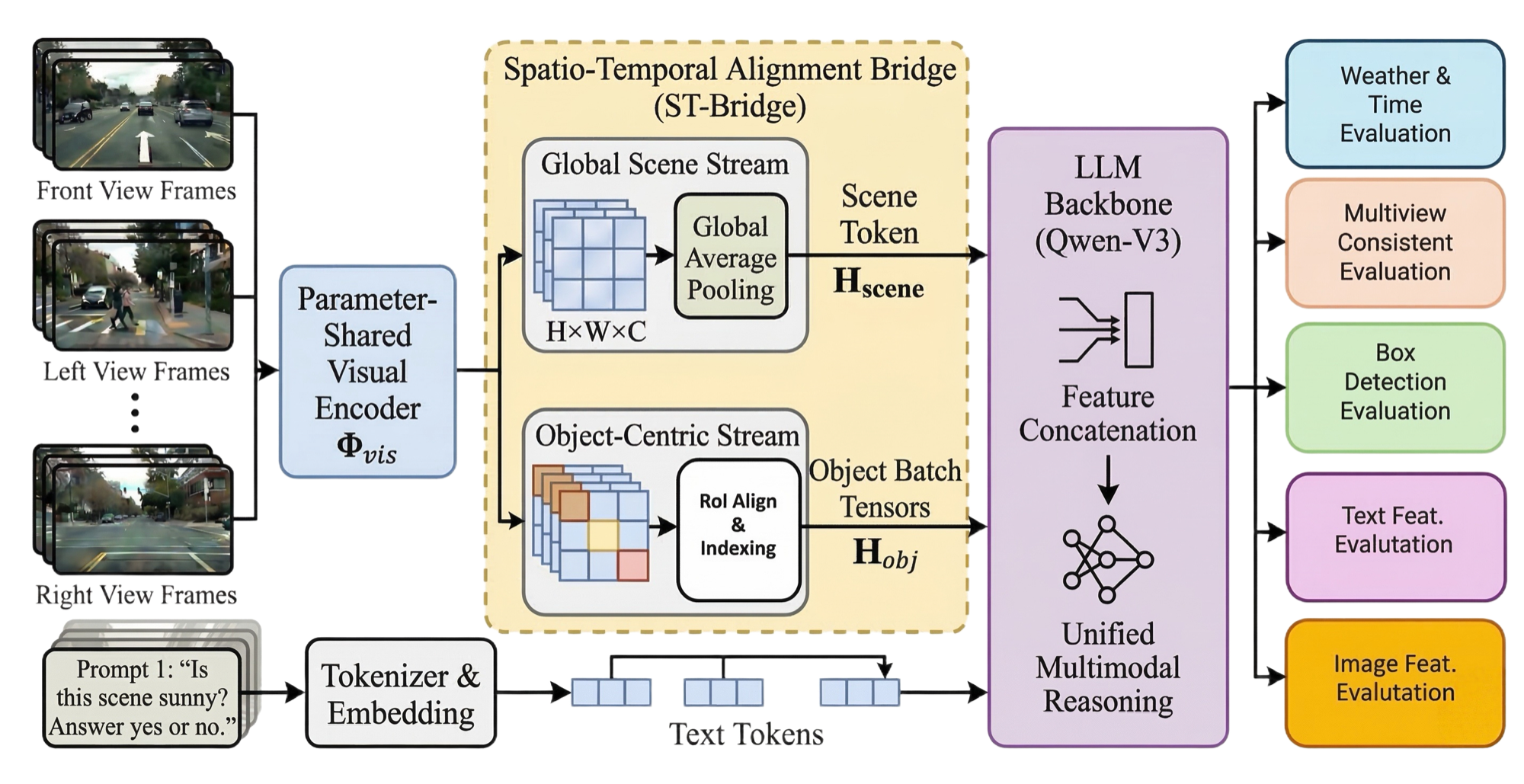}
    \caption{\textbf{The illustration of MV-VLM structure.} }  
    \label{fig:mv-vlm structure}
\end{figure}

\textbf{MV-VLM Structure.}
We instantiate the intelligent evaluator $\mathcal{E}$ using a "Dual-Stream Perception, Unified Reasoning" paradigm built upon the Qwen-V3 \cite{qwen3} architecture. As illustrated in Fig. \ref{fig:mv-vlm structure}, the model processes multi-view video frames through a parameter-shared visual encoder $\Phi_{vis}$. To bridge the modality gap, we introduce a {Spatio-Temporal Alignment Bridge (ST-Bridge)} comprising two parallel branches: (1) a \textit{Global Scene Stream} that compresses holistic features into scene tokens $\mathbf{H}_{scene}$ for macro-level auditing; and (2) an \textit{Object-Centric Stream} that utilizes index-guided RoI alignment \cite{yang2024semantic} to aggregate specific object instances across time and views into batch tensors $\mathbf{H}_{obj}$. These hierarchical visual tokens are concatenated with textual instructions and fed into the LLM backbone, enabling unified evaluation of both global consistency and fine-grained object identity preservation. \clh{Specifically, the unified multimodal reasoning module outputs a comprehensive suite of assessments across five key \ssh{evaluations}: global weather and time accuracy, multiview spatiotemporal consistency, bounding box detection precision, as well as text and image feature alignment. These multi-dimensional outputs serve as explicit diagnostic signals to determine whether the generated frames pass the quality audit or require targeted regeneration and refinement \ssh{as below}.}

\textbf{Macro-level Scene Assessment (Regeneration Loop).} The agent first evaluates the global consistency of the generated video $\hat{\mathbf{x}}$ against the global conditions $\mathbf{c}_{global}$ (e.g., weather, time of day) using a macro-consistency score $S_{macro}$ calculated as:\begin{equation}S_{macro} = \mathcal{E}(\hat{\mathbf{x}}, \mathbf{c}_{global}).\end{equation} If $S_{macro}$ falls below a predefined threshold $\gamma_{g}$, the system identifies a "Major Flaw." In this case, we reinforce the constraint weights, emphasizing neglected attributes (e.g., "heavy rain" or "midnight"), to form an enhanced condition $\mathbf{c}'_{global}$ and trigger a full regeneration:\begin{equation}\mathbf{c}'_{global} = \text{Emphasize}(\mathbf{c}_{global}), \quad \hat{\mathbf{x}}_{new} \leftarrow \mathcal{G}(\mathbf{z}_{init}, \mathbf{c}'_{global}).\end{equation}

\textbf{Micro-level Object Assessment (Refinement Loop).} Upon passing the macro-level check, the agent proceeds to fine-grained object evaluation. To robustly assess the identity preservation and visual quality of specific agents, we propose a Multi-View Batch Evaluation strategy. For the $i$-th object in the scene, we crop and aggregate its visual instances across all $N$ camera views and $T$ frames into a unified batch tensor $\mathcal{B}_i$:
\begin{equation}
\mathcal{B}_i = { \text{Crop}(\hat{\mathbf{x}}_{t}^{v}, b_{i,t}^{v}) \mid v \in [1, N], t \in [1, T] ,}\end{equation}
where $b_{i,t}^{v}$ denotes the bounding box of object $i$ at frame $t$ in view $v$. 
\clh{Based on this aggregated batch, we compute the object-level consistency score $S_{obj}^{i}$, which comprehensively evaluates both \textit{semantic alignment} and \textit{visual clarity}.}


\clh{
To resolve the dimensionality mismatch between the VLM's visual feature space and the fused multimodal condition $\mathbf{e}_i$, we introduce projection heads $\phi_{v}(\cdot)$ and $\phi_{e}(\cdot)$ to map both representations into a shared semantic latent space. Furthermore, to ensure the metrics are evaluated on a consistent scale, we apply $\ell_2$-normalization to the projected features prior to similarity computation. The comprehensive score is formulated as:
\begin{equation}
\label{eq:S_obj}
    S_{obj}^{i} = \lambda \left( \frac{\phi_v(\mathcal{E}_{vis}(\mathcal{B}_i)) \cdot \phi_e(\mathbf{e}_i)}{\|\phi_v(\mathcal{E}_{vis}(\mathcal{B}_i))\|_2 \|\phi_e(\mathbf{e}_i)\|_2} \right) + (1 - \lambda) \widetilde{\mathcal{Q}}_{clarity}(\mathcal{B}_i),
\end{equation}
where $\mathcal{E}_{vis}$ is the visual encoder of the VLM, and $\widetilde{\mathcal{Q}}_{clarity} \in [0, 1]$ is the min-max normalized perceptual quality score measuring artifacts (e.g., sharpness and lack of distortion). The hyperparameter $\lambda \in (0, 1)$ balances the semantic alignment and visual clarity criteria. If $S_{obj}^{i}$ falls below a predefined satisfactory threshold (indicating a "Minor Flaw"), the agent flags this specific object and invokes the refinement module (detailed in Sec. \ref{sec:obj finetune}). This targeted refinement repairs inconsistent regions while preserving the background, ultimately yielding a high-quality, long-sequence, multiview, spatial-temporal-consistent video that meets user expectations.
}
\begin{figure}[t]
    \centering
    \includegraphics[width=\linewidth]{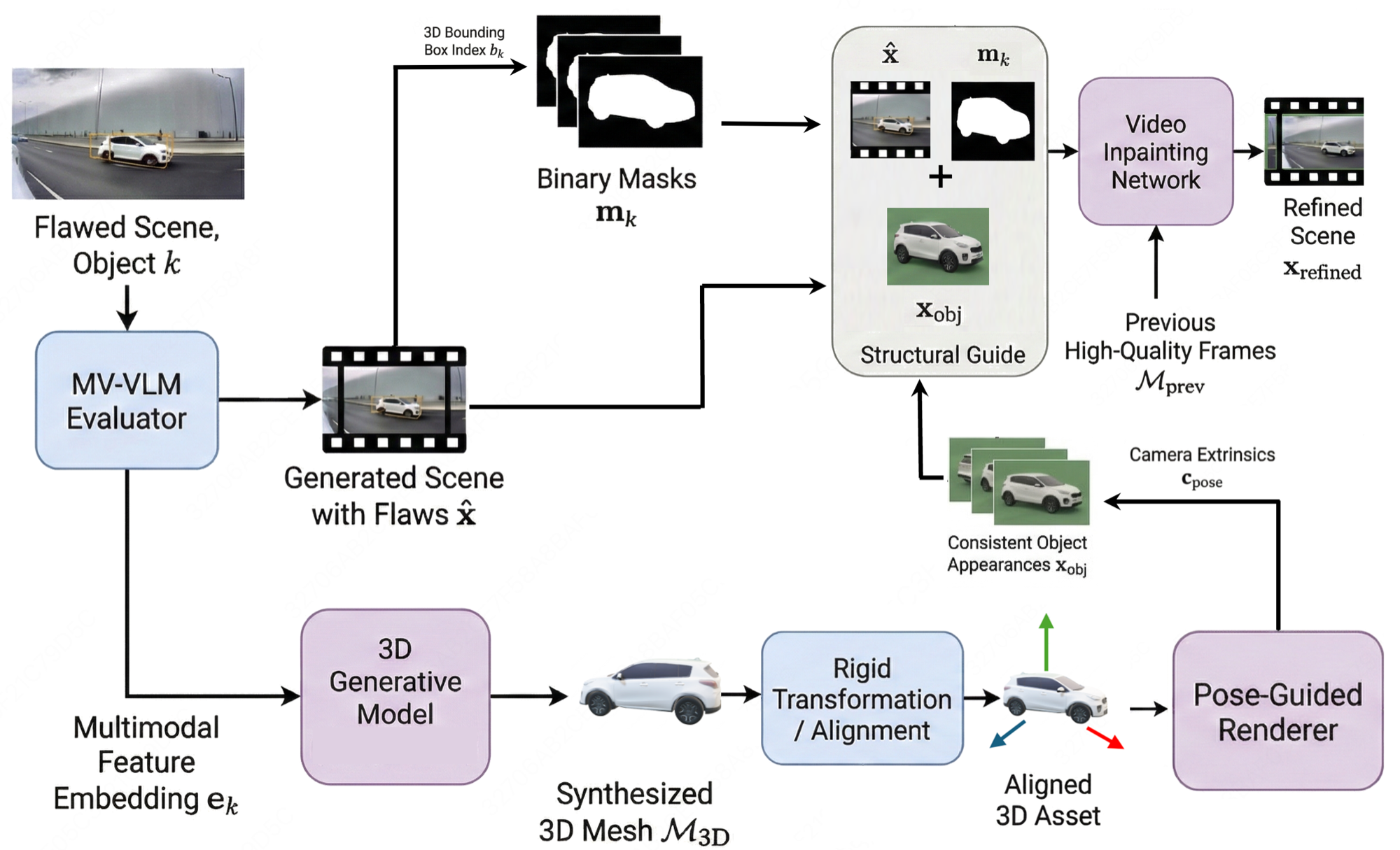}
    \caption{\textbf{The workflow of our object-level refinement module.} }  
    \label{fig:refine}
\end{figure}

\vspace{-0.5em}
\subsection{Object-Level Refinement}\label{sec:obj finetune}

We further propose an object-level refinement module $\mathcal{R}$ (as shown in Fig.~\ref{fig:refine}) that leverages explicit 3D geometric cues to ensure intrinsic spatiotemporal coherence.

\textbf{Fault Localization and Extraction.}
The refinement process is triggered when the MV-VLM Evaluator flags a specific object $k$ as inconsistent. The evaluator provides the object's unique 3D bounding box index $b_k$ and its aligned multimodal feature embedding $\mathbf{e}_k$ (as defined in Eq. 5).
Using the spatio-temporal trajectory defined by $b_k$, we first generate a sequence of 3D-projected binary masks $\mathbf{m}_k$ to isolate the flawed region from the generated scene $\hat{\mathbf{x}}$. This precise localization ensures that the refinement is strictly confined to the target object without altering the coherent background.

\textbf{3D Asset Synthesis and Embedding.}
To fundamentally resolve cross-view inconsistencies, we move beyond 2D pixel-level repair and reconstruct the object in 3D. We utilize a state-of-the-art 3D generative model \cite{zhao2025hunyuan3d} conditioned on the feature description $\mathbf{e}_k$:
\begin{equation}
\mathcal{M}_{3D} = \text{Gen}_{3D}(\mathbf{e}_k),
\end{equation}
where $\mathcal{M}_{3D}$ represents the synthesized 3D mesh 
that perfectly matches the semantic description, aligned with the bounding box trajectory $b_k$ within the scene's global coordinate system.

\textbf{Pose-Guided Rendering and Fusion.}
With the consistent 3D model $\mathcal{M}_{3D}$ placed in the virtual environment, we render it onto the 2D image planes using the camera extrinsics $c_{pose}$ from the original multi-view setup. This process yields a sequence of consistent object appearances $\mathbf{x}_{obj}$.
Finally, to ensure seamless blending with the lighting and shadows of the original scene, we employ a video inpainting network.\cite{liang2025driveeditor} This network takes the rendered 3D object $\mathbf{x}_{obj}$ as a strong structural guide (Control Signal) and the previous high-quality frames $\mathcal{M}_{prev}$ as temporal context:
\begin{equation}
\mathbf{x}_{refined} = \text{Inpaint}(\hat{\mathbf{x}} \odot (1-\mathbf{m}_k) + \mathbf{x}_{obj} \odot \mathbf{m}_k, \mathcal{M}_{prev}).
\end{equation}

%% file: sec/3-exp.tex
\section{Experiments}

\subsection{Implementation Details}
\textbf{Backbones and Training.} We use the generative backbone of video generator from~\cite{gao2025magicdrive}, which 
perform the diffusion process from Gaussian noise $\mathbf{z}_{init}$ by solving an Ordinary Differential Equation (ODE). using core network $v_\theta$ (MV-STDiT) processes multi-view latents through a hierarchy of Spatial-Temporal Attention Blocks. To ensure consistency across views and coherence over time, we integrate \textit{Cross-View Attention} and \textit{Temporal Attention} mechanisms into each block. 

\begin{figure*}[htbp]
    \centering
    \includegraphics[width=0.9\linewidth]{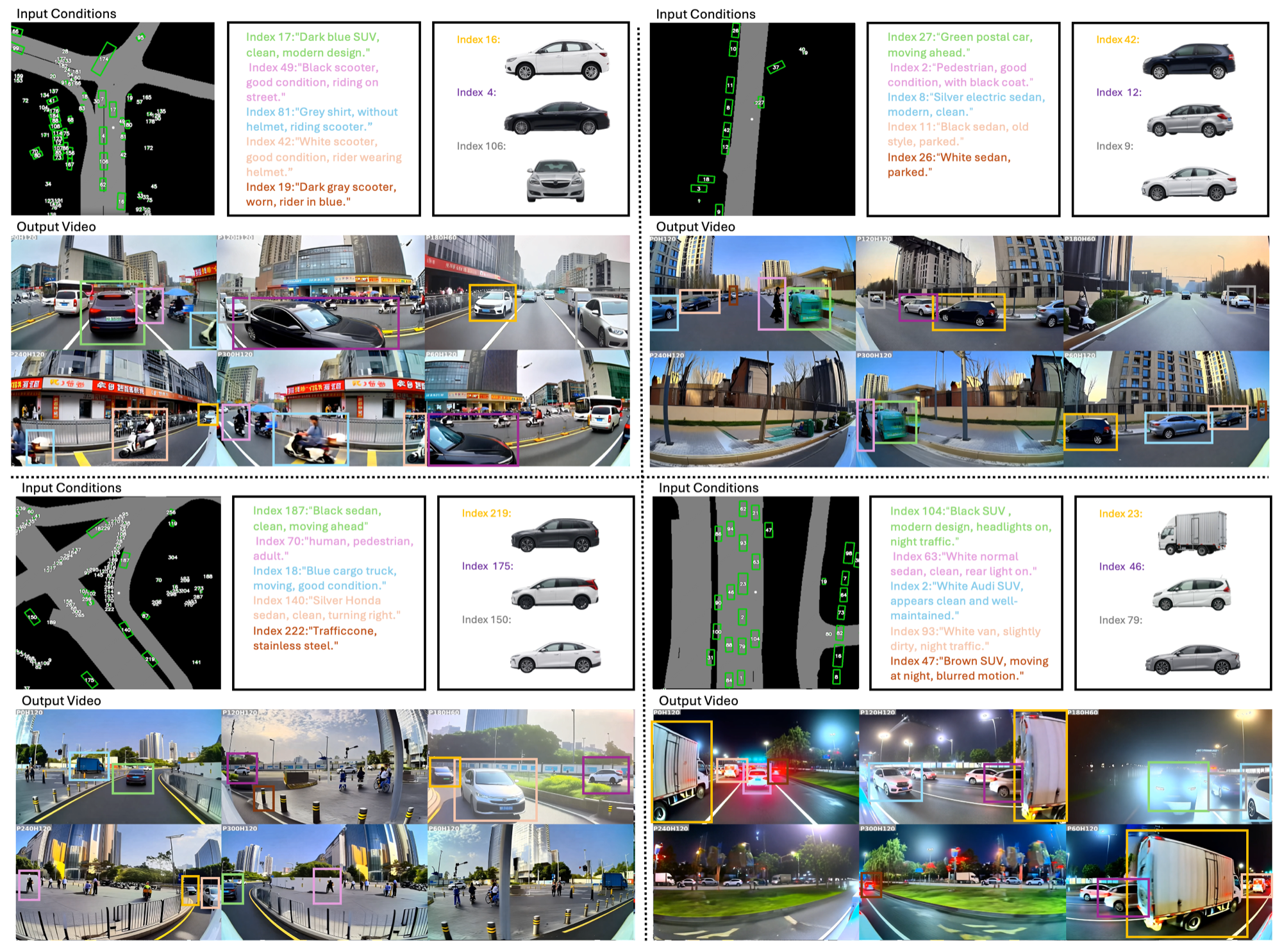}
    \caption{\clh{\textbf{Qualitative results of fine-grained video generation by VistaGEN in complex scenes.} Our model seamlessly integrates multimodal conditions, including BEV layouts, textual descriptions, and reference images. The colored bounding boxes track specific controlled instances across the generated multi-view videos, demonstrating precise object-level controllability.}} 
    \label{fig:our_fig2}
    
\end{figure*}

\textbf{Comparing Baselines.}
To evaluate the effectiveness of our proposed framework, we benchmark it against two state-of-the-art driving video generation models: \textbf{MagicDrive-V2} \cite{gao2025magicdrive}, representing the leading DiT-based approach with geometric control; and \textbf{DriveDreamer-2} \cite{zhao2025drivedreamer}, a prominent world model known for long-sequence synthesis. These baselines provide a comprehensive reference for both geometry-aware generation and temporal coherence.

\textbf{Datasets.} We train our model on a self-collected large-scale proprietary dataset, which shares a data structure similar to nuScenes \cite{caesar2020nuscenes} but is distinctively enriched with comprehensive object-level textual descriptions. \clh{Collected from real-world road environments, the dataset encompasses a wide spectrum of driving conditions, including various complex traffic scenarios. The training set comprises 40,000 multi-view video clips, where each clip consists of 97 synchronized frames spanning a duration of 12 seconds.}
\clh{For evaluation, we utilize a strictly disjoint validation set of 4,000 cases. These cases were selected via [random sampling / chronologically / from distinct driving logs], ensuring strictly no overlap with the training set to prevent data leakage. To further enhance the model's generalization capabilities, we fine-tune our pre-trained model on the public nuScenes dataset \cite{caesar2020nuscenes}. This step allows us to rigorously assess the model's scalability and its ability to generate diverse environmental contexts.}

\textbf{Evaluation Metrics.} 
We adopt a multi-dimensional protocol to assess performance across three key aspects:
First, to evaluate \textbf{Video Quality and Layout Fidelity}~\cite{chen2025eccv}, we report \textbf{FID} and \textbf{FVD} for perceptual realism, while using \textbf{mAP} and \textbf{IoU} to measure the precision of generated agents against input bounding box layouts.
Moreover, to verify our \textbf{Fine-Grained Controllability} and \textbf{Index Consistency}, we employ VLM evaluation, which measures the semantic and visual alignment of generated objects with their respective text descriptions and reference images.

\begin{table}[tbp]
    \centering
    \caption{\textbf{Comparison of Video Quality and Layout Control.} We evaluate perceptual quality using FVD/FID and geometric precision using mAP/IoU. \clh{Assessments are conducted on the validation sets of both the nuScenes dataset and our self-collected dataset.} $\downarrow$ indicates lower is better, $\uparrow$ indicates higher is better. 
    }
    \label{tab:video_quality}
    \resizebox{\linewidth}{!}{
    \begin{tabular}{l cc cc} 
        \toprule
        \multirow{2}{*}{\textbf{Method}} & \multicolumn{2}{c}{\textbf{Perceptual Quality}} & \multicolumn{2}{c}{\textbf{Layout Fidelity}} \\
        \cmidrule(lr){2-3} \cmidrule(lr){4-5} 
         & \textbf{FVD} ($\downarrow$) & \textbf{FID} ($\downarrow$) & \textbf{mAP} ($\uparrow$) & \textbf{IoU} ($\uparrow$) \\
        \midrule
        BEVControl \cite{yang2023bevcontrol} & 112.3 & 12.01 & N/A & 52.37 \\ 

        MagicDrive-V2 \cite{gao2025magicdrive} & \textbf{105.2} & 11.54 & 43.50 & 53.80 \\
        
        DriveDreamer \cite{wang2024drivedreamer} & 113.8 & 11.85 & 38.45 & 47.20 \\
        DriveDreamer-2 \cite{zhao2025drivedreamer} & 110.3 & 11.20 & 41.30 & 50.10 \\
        \midrule
        
        \textbf{Ours} & 107.7 & \textbf{11.03} & \textbf{53.48} & \textbf{65.70} \\
        \bottomrule
    \end{tabular}
    }
\end{table}

\subsection{Generation Quality Evaluation}
\label{sec:generation quality}

We first conduct experiments to quantitatively evaluate the driving video generation quality of our VistaGEN, including video quality and layout fidelity, spatiotemporal consistency, and controllability, and compare it with state-of-the-art driving video generation approaches~\cite{yang2023bevcontrol,gao2025magicdrive,wang2024drivedreamer,zhao2025drivedreamer}, followed by a qualitative evaluation. 

\textbf{Quantitative Evaluation.}
We conduct a comprehensive quantitative comparison against state-of-the-art baselines, including \clh{BEVControl,} MagicDrive-V2, DriveDreamer and DriveDreamer-2. The results are categorized into general video quality/layout control (Table \ref{tab:video_quality}) and spatio-temporal consistency/fine-grained alignment (Table \ref{tab:vlm_evaluation}).

\textbf{Video Quality and Layout Fidelity.}
As presented in Table \ref{tab:video_quality}, our proposed framework maintains a comparable level of perceptual quality and layout adherence to the strong baseline, MagicDrive-V2.
Specifically, we achieve an FVD of \textbf{107.7} and an FID of \textbf{11.03}, which are on par with MagicDrive-V2. This demonstrates that introducing our fine-grained object control mechanism does not compromise global generation fidelity, a balance achieved by our robust DiT backbone and progressive training strategy.
Regarding geometric control, our method also remains highly competitive, achieving an mAP of \textbf{53.48} and an IoU of \textbf{65.70}. This confirms that our hierarchical condition injection effectively preserves the precise spatial arrangement of dynamic agents, minimizing layout drift even in challenging long sequences.

\begin{table*}[htbp]
    \centering
    \caption{\textbf{Quantitative Evaluation using MV-VLM Evaluator.} We report the accuracy and consistency scores, where "Real Data" serves as the reference upper bound. 
    }
    \label{tab:vlm_evaluation}
    \resizebox{0.95\linewidth}{!}{
    \begin{tabular}{l|cccc|cc}
        \toprule
        \multirow{2}{*}{\textbf{Metric}} & \textbf{BEVControl} & \textbf{MagicDrive-V2} & \textbf{DriveDreamer} & \textbf{DriveDreamer-2} & \textbf{Ours} & \textbf{Real Data (Ref)} \\
         & \cite{yang2023bevcontrol} & \cite{gao2025magicdrive} & \cite{wang2024drivedreamer} & \cite{zhao2025drivedreamer} & \textbf{(Generated)} & \textbf{(Oracle)} \\
        \midrule
        \multicolumn{7}{l}{\textit{Global Scene Level}} \\
        Weather Accuracy ($\uparrow$) & 73.2\% & 75.4\% & 72.8\% & 74.3\% & \textbf{75.6\%} & 76.6\% \\
        Time Accuracy ($\uparrow$)    & 75.14\% & 77.39\% & 76.50\% & 78.31\% & \textbf{80.9\%} & 83.0\% \\
        \midrule
        \multicolumn{7}{l}{\textit{Object Level (Box-Guided)}} \\
        Category Accuracy ($\uparrow$) & 76.5\% & 79.9\% & 75.9\% & 78.3\% & \textbf{85.1\%} & 88.5\% \\
        Image Alignment (SigLIP) ($\uparrow$) & 8.5\% & 10.2\% & 9.8\% & 11.3\% & \textbf{70.7\%} & 88.2\% \\
        Text Alignment ($\uparrow$)    & 12.1\% & 15.4\% & 12.5\% & 14.7\% & \textbf{60.3\%} & 73.1\% \\
        \midrule
        \multicolumn{7}{l}{\textit{Spatio-Temporal Consistency}} \\
        \textbf{Index Consistency} ($\uparrow$) & 68.4\% & 72.3\% & 67.5\% & 70.1\% & \textbf{84.9\%} & 93.6\% \\
        \bottomrule
    \end{tabular}
    }
\end{table*}

\begin{figure*}[htbp]
    \centering
    \includegraphics[width=\linewidth]{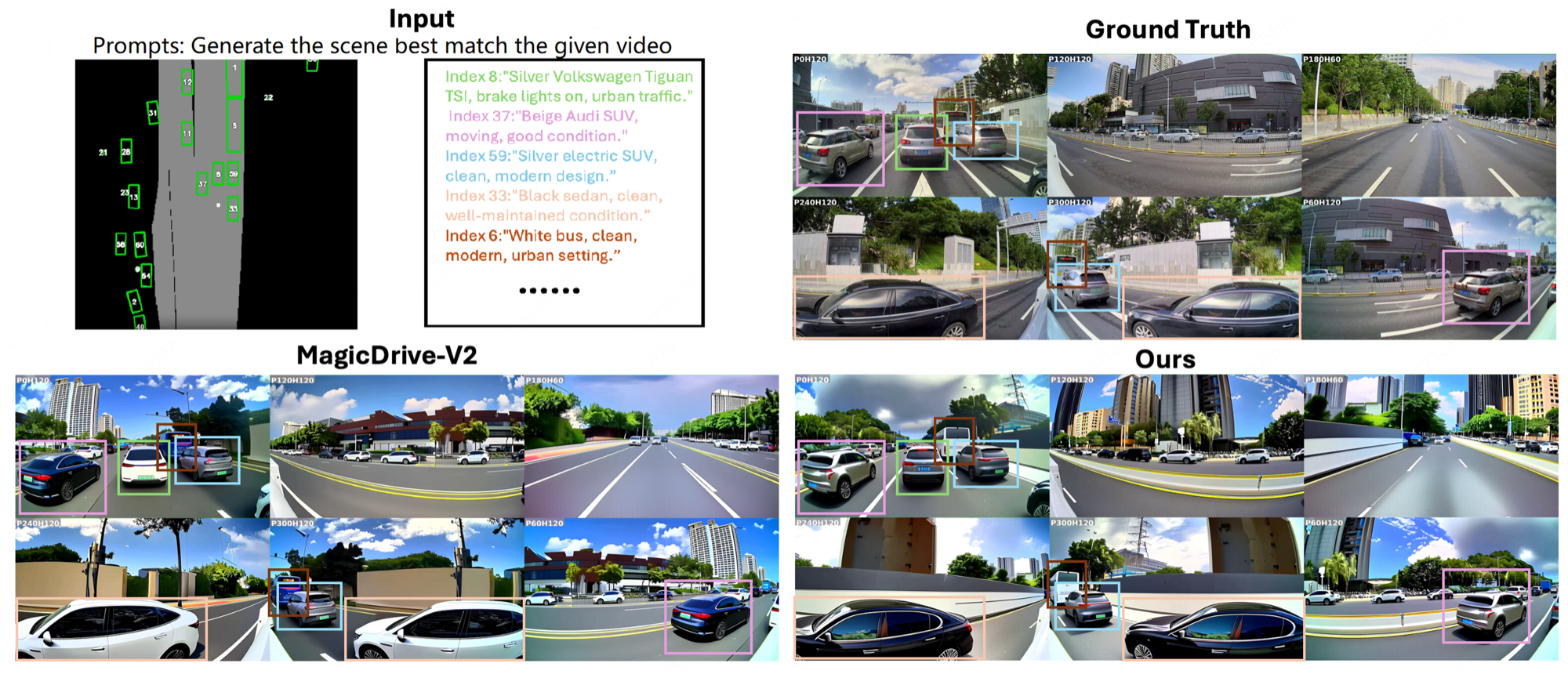}
    \caption{Qualitative video generation comparison results evaluated on our self-collected real-world dataset from different comparing approaches, using prompt: Generating the same scene and objects as GT. The GT reference images are also demonstrated.}  
    \label{fig:exp_ourdataset_1}

\end{figure*}

\begin{figure*}[htbp]
    \centering
    \includegraphics[width=\linewidth]{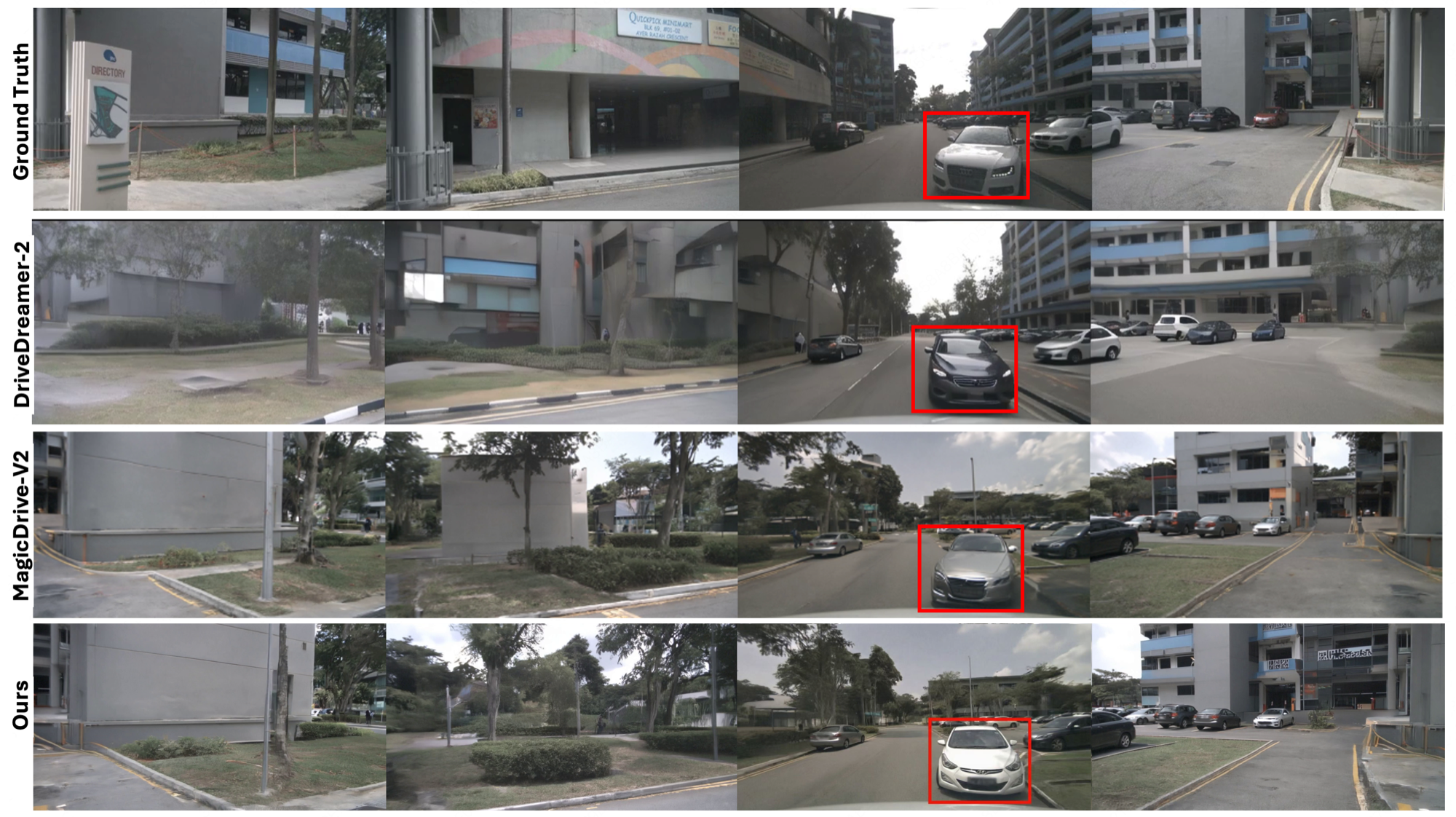}
    \caption{Qualitative video generation comparison results evaluated on nuScenes dataset from different comparing approaches, including DriveDreamer-2~\cite{zhao2025drivedreamer}, MagicDrive-V2~\cite{gao2025magicdrive} and ours, using the same text prompt: A white sedan, clean, facing camera. The GT reference images are also demonstrated. }  
    \label{fig:exp_nuScenes_1}
\end{figure*}

\begin{figure*}[htbp]
    \centering
    \includegraphics[width=\linewidth]{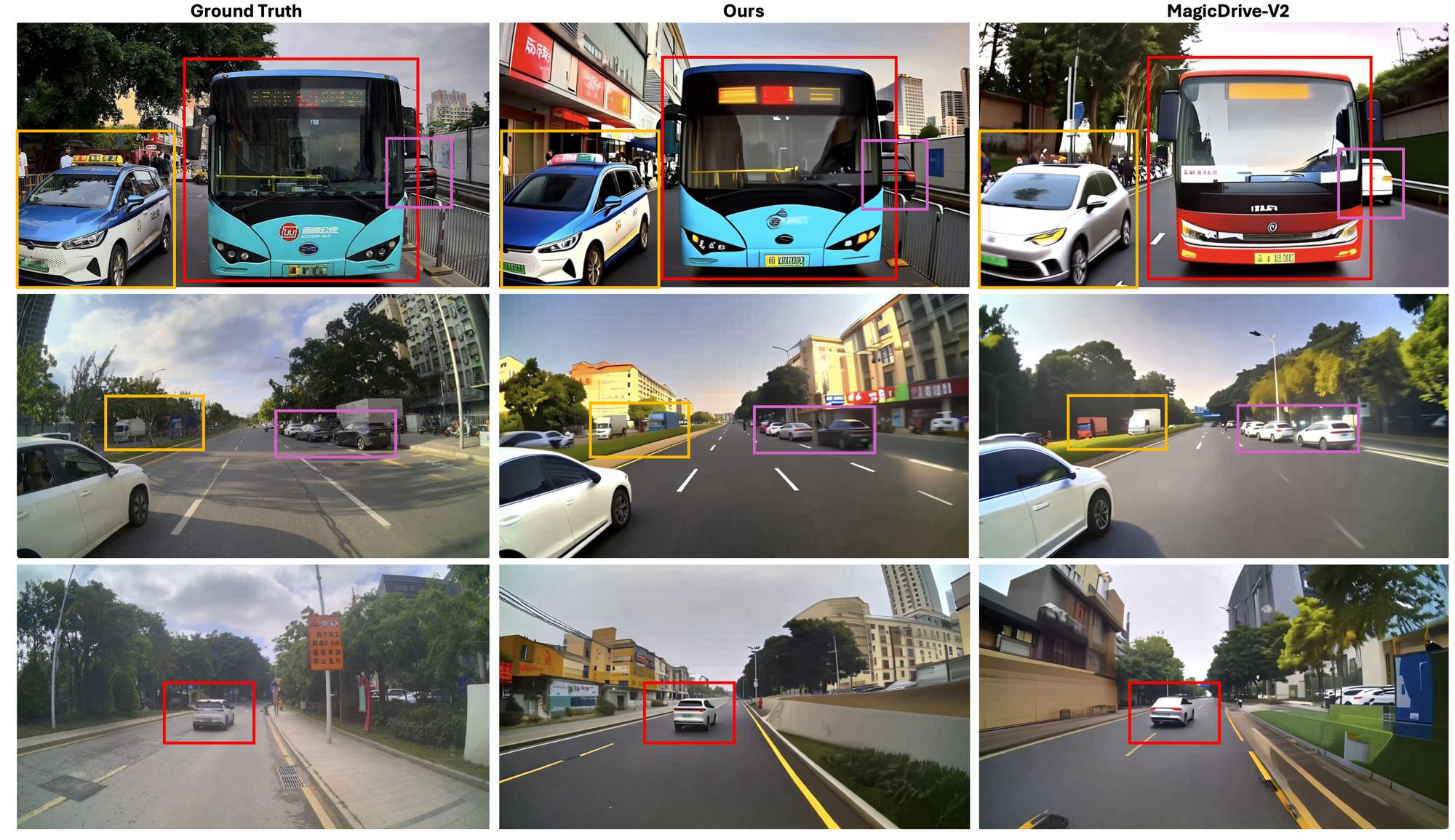}
    \caption{Qualitative video generation comparison results evaluated on our dataset from different comparing approaches, including MagicDrive-V2~\cite{gao2025magicdrive} and ours, using the same GT input video.}  
    \label{fig:exp_ourdataset_compared}
\end{figure*}

\textbf{Qualitative Evaluation.} 
We also show visual comparison results for different input conditions across the compared approaches. \ssh{For an efficient comparison, we choose MagicDrive-V2~\cite{gao2025magicdrive} and DriveDreamer-2~\cite{zhao2025drivedreamer} as baselines since these two approaches are state-of-the-art driving video generation methods.} As we can see in Fig.~\ref{fig:our_fig2}, our VistaGEN can achieve 
\clh{fine-grained control generation with specific text prompts.} As seen in Fig.~\ref{fig:exp_ourdataset_1}, Fig. \ref{fig:exp_nuScenes_1} and Fig. \ref{fig:exp_ourdataset_compared}, \textbf{VistaGen} exhibits superior alignment with Ground Truth (GT) inputs compared to these two baseline methods.

\textbf{MV-VLM Evaluation.} To quantitatively assess the generation quality from the perspective of a general-purpose vision-language model, we calculate the following metrics: (1)\textbf{Scene Attributes Accuracy,} the consistency of global environmental conditions, 
(2)\textbf{Object Category Accuracy,} the generated objects at specified bounding box locations match the requested categories (e.g., car, bus, pedestrian),(3)\textbf{Visual \& Semantic Alignment,} including \textit{Image Alignment} (SigLIP score) 
and \textit{Text Alignment}, 
and (4) \textbf{Index Consistency,} measuring whether the same object (tracked by Index ID) maintains a consistent visual identity across multiple frames and camera views.

Table \ref{tab:vlm_evaluation} presents the evaluation results. 
Our VistaGEN demonstrates exceptional performance, particularly in \textbf{Category Accuracy (85.1\%)} and \textbf{Index Consistency (84.9\%)}, closely approaching the oracle performance of real data (88.5\% and 93.6\%, respectively). This indicates that our method successfully generates stable, recognizable objects that adhere to physical layouts.
While \textbf{Text} and \textbf{Image Alignment} scores (60.3\% and 70.7\%) show a gap compared to real data—reflecting the inherent difficulty of fine-grained attribute injection—our method significantly outperforms baselines without explicit object-level control mechanisms.

\begin{figure}[htbp]
    \centering
    \includegraphics[width=\linewidth]{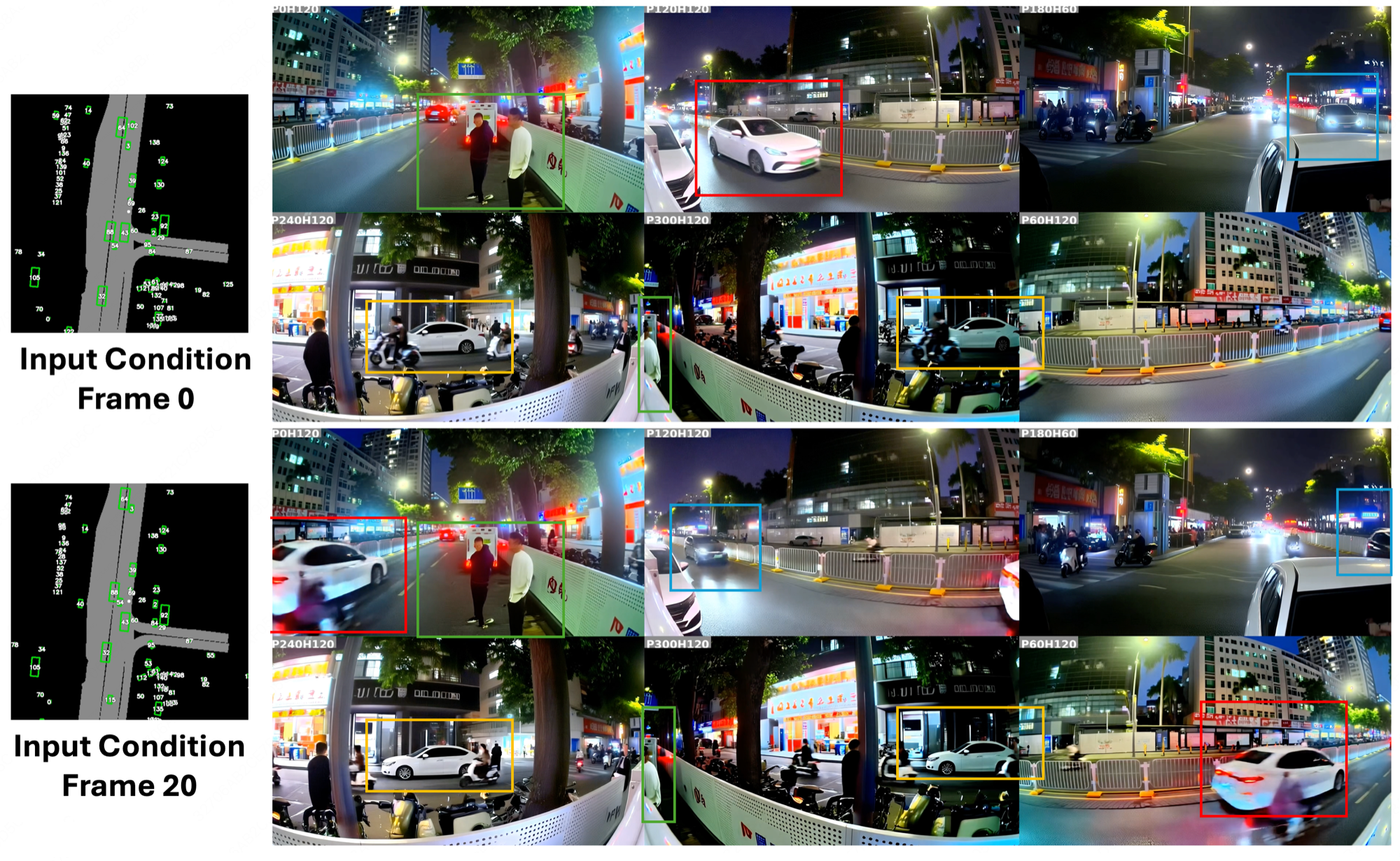}
    \caption{More visual generation results by our VistaGEN, which preserve multiview spatial-temporal consistency in complex scene.}  
    \label{fig:complex_scene_1}
\end{figure}

\ssh{\subsection{Generation Controllability Evaluation}}
\clh{
\textbf{Spatio-Temporal Consistency and Controllability.}
Table~\ref{tab:vlm_evaluation} presents a quantitative comparison of spatiotemporal consistency and fine-grained controllability against baseline methods: 
(1) \textit{Consistency:} Our method achieves superior Index Consistency accuracy. While baselines frequently suffer from object flickering or cross-view deformation in complex scenes, our object-level refinement module—guided by the 3D proxy—successfully rectifies these artifacts, ensuring rigid 3D consistency across multi-camera views. 
(2) \textit{Fine-Grained Controllability:} In terms of attribute alignment, our model demonstrates a substantial lead with a \clh{Text Alignment accuracy} of \textbf{60.3\%} and an \clh{Image Alignment accuracy} of \textbf{70.7\%}. This confirms that our multimodal condition injection mechanism effectively translates fine-grained textual and visual prompts into the generated video, a capability that previous coarse-grained control methods lack. 
As visually corroborated in Fig.~\ref{fig:complex_scene_1}, these quantitative advantages seamlessly translate into high-fidelity generation results, where VistaGEN preserves strict multiview spatiotemporal consistency and precise object attributes even in highly complex environments.
}

\begin{figure*}[htbp]
    \centering
    \includegraphics[width=\linewidth]{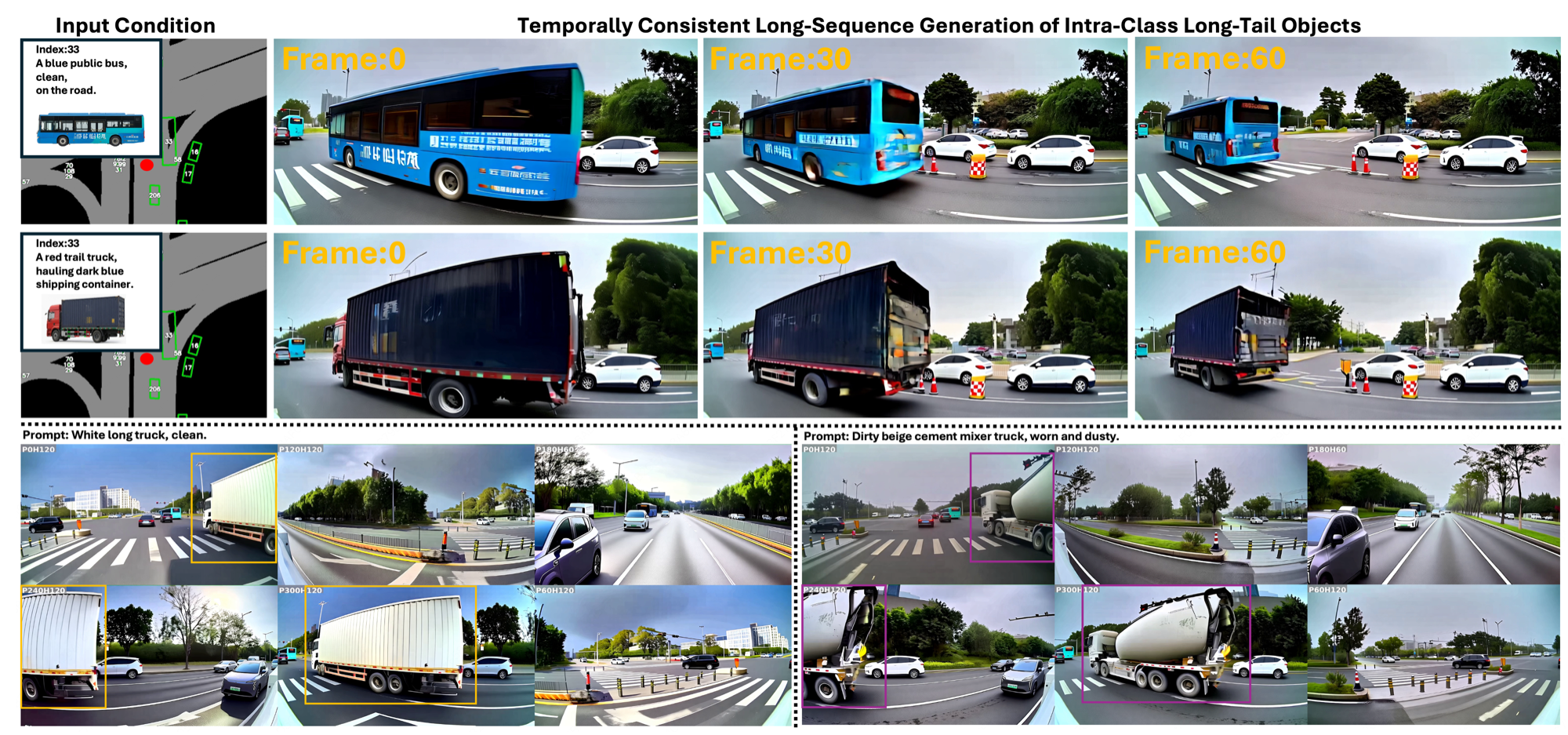}
    \caption{More visual results of VistaGEN using different prompts. {Our VistaGEN can easily achieve intra-class long-tail generation with reasonable and coherent generation given different prompts while maintaining spatiotemporal consistency.} }  
    \label{fig:long_tail}
\end{figure*}

\clh{
\textbf{Intra-Class Long-Tail Generation.}
Building upon the superior fine-grained controllability, VistaGEN exhibits exceptional capability in addressing the intra-class long-tail generation problem. Standard autonomous driving datasets are typically dominated by common vehicle types (e.g., standard sedans or SUVs), making it notoriously difficult for conventional models to synthesize rare instances such as construction vehicles, cement trucks, or trailers. However, driven by our precise local condition injection and the robust closed-loop refinement, our model can generalize well to these under-represented categories. As visually demonstrated in Fig. \ref{fig:long_tail}, VistaGEN successfully synthesizes complex, atypical construction vehicles under varying textual prompts and achieves spatiotemporal consistency simultaneously. The generated multiview sequences maintain high structural fidelity and coherent semantic details, proving that our framework can effectively mine and generate rare data to enrich autonomous driving simulations.}

\textbf{Comparison with General Video Generation Approaches.}
We also compare with some general video generation models, including SOTA general-purpose foundation models (e.g., Wan2.1 \cite{wan2025}, Cosmos2.5 \cite{ali2025world}) and other domain-specific methods (e.g., MagicDrive \cite{gao2023magicdrive}, StreetScape \cite{deng2024streetscapes}).
Here, we focus only on fine-grained controllability and spatiotemporal consistency during the comparison. Compared with our VistaGEN, these approaches lack a fine-grained conditioning interface. Without explicit multi-view geometric constraints, they also suffer from severe degradation in spatiotemporal consistency when generating long sequences across multiple camera views. 
For a more detailed visual analysis and comparison of these failure cases, please refer to our {supplementary materials}.

\subsection{Ablation Studies}
\begin{figure}[t]
    \centering
    \includegraphics[width=\linewidth]{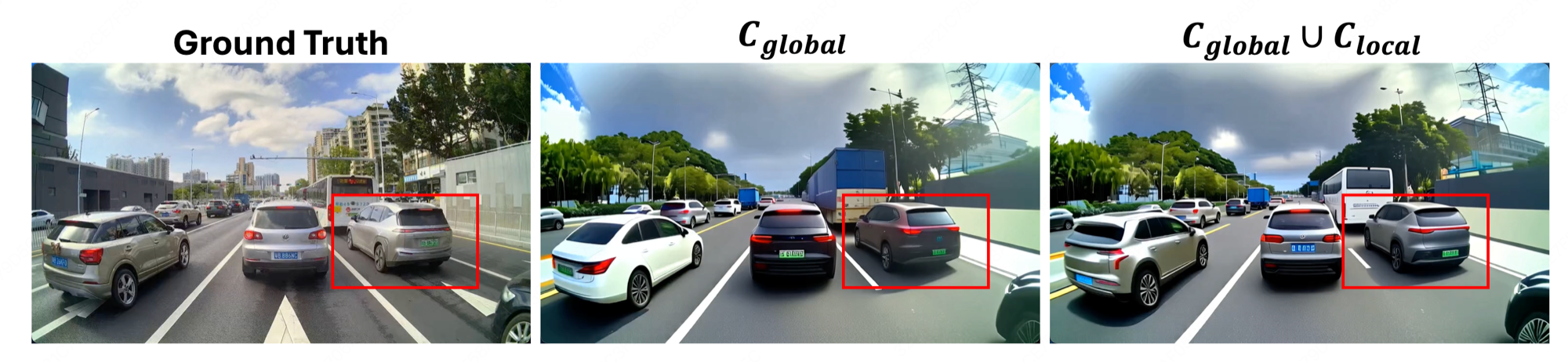}
    \caption{ Visual comparison results without (middle) or with (right) using the visual-language feature condition. }  
    \label{fig:ablation_condition}
    \vspace{-0.5cm}
\end{figure}

\textbf{Visual-Language Feature Condition.}
We investigate the contribution of the visual-language based local object condition $\mathbf{c}_{local}$ to the final generation with two settings: (1) using only the global context $\mathbf{c}_{global}$, and (2) enabling both global and local conditions ($\mathbf{c}_{global} + \mathbf{c}_{local}$). As shown in Table \ref{tab:ablation_condition}, removing $\mathbf{c}_{local}$ leads to a significant drop in fine-grained alignment metrics. While the geometric precision (mAP) remains stable—since the bounding boxes are provided in both settings—the 
Text Alignment accuracy and Image Alignment decrease sharply. This indicates that without local guidance, the model generates generic objects that fail to match specific semantic attributes. We also show some visual comparison results in Fig. \ref{fig:ablation_condition}, where the global condition specifies "a clear sunny day, morning," while the local condition for the target vehicle is "{Silver electric SUV, clean, modern style.}" With the injection of $\mathbf{c}_{local}$, our VistaGEN successfully steers the generation to the description, producing a vehicle with the correct silver color and modern body type.

\clh{
\textbf{Effectiveness of the MV-VLM Evaluator.}
A core contribution of VistaGEN is the transition from an open-loop, single-pass generation to a closed-loop paradigm equipped with evaluation feedback. To quantify this contribution, we conduct an ablation study comparing the full system against a "Direct Generation" baseline, where the multiview video is produced in a single forward pass without the MV-VLM evaluator or the subsequent refinement module. 
As shown in Table \ref{tab:ablation_mvvlm}, our proposed closed-loop paradigm yields substantial improvements across all metrics. While direct generation struggles with visual artifacts and precise control, our full system effectively enhances perceptual video quality (improving both FVD and FID). More importantly, the closed-loop mechanism significantly boosts layout fidelity (mAP and IoU) and fine-grained semantic alignment (CLIP-T and CLIP-I). This confirms that the evaluation-refinement feedback loop is crucial for enforcing strict geometric constraints and accurate multimodal attribute injection throughout the generated sequences.
}

\begin{table}[h]
    \centering
    \caption{\textbf{Ablation on Object-Level Conditioning.} We compare multiview video generation performance with and without the local object prompt $\mathbf{c}_{local}$. While geometric accuracy (mAP) remains high in both settings due to box constraints, semantic alignment improves significantly with $\mathbf{c}_{local}$.}
    \label{tab:ablation_condition}
    \resizebox{\linewidth}{!}{
    \begin{tabular}{l|c|cc}
        \toprule
        \textbf{Condition Setting} & \textbf{Layout (mAP) $\uparrow$} & \textbf{Text Alignment $\uparrow$} & \textbf{Image Alignment $\uparrow$} \\
        \midrule
        Global Only ($\mathbf{c}_{global}$) & 53.03 & 15.7 & 17.4 \\
        Global + Local ($\mathbf{c}_{global} + \mathbf{c}_{local}$) & \textbf{53.48} & \textbf{50.5} & \textbf{57.1} \\
        \bottomrule
    \end{tabular}
    }
\end{table}

\begin{table}[h]
    \centering
    \caption{\textbf{Effectiveness of the MV-VLM Evaluator.} Qualitative comparison between direct open-loop generation and our proposed closed-loop paradigm.}
    \label{tab:ablation_mvvlm}
    \resizebox{\linewidth}{!}{
    \begin{tabular}{l cc cc cc}
        \toprule
        \multirow{2}{*}{\textbf{Settings}} & \multicolumn{2}{c}{\textbf{Video Quality}} & \multicolumn{2}{c}{\textbf{Layout Fidelity}} & \multicolumn{2}{c}{\textbf{Alignment}} \\
        \cmidrule(lr){2-3} \cmidrule(lr){4-5} \cmidrule(lr){6-7}
         & \textbf{FVD} ($\downarrow$) & \textbf{FID} ($\downarrow$) & \textbf{mAP} ($\uparrow$) & \textbf{IoU} ($\uparrow$) & \textbf{CLIP-T} ($\uparrow$) & \textbf{CLIP-I} ($\uparrow$) \\
        \midrule
        Direct Generation & 110.4 & 11.53 & 45.52 & 555.73 & 0.667 & 0.686 \\
        \textbf{Closed-Loop Generation} & \textbf{107.7} & \textbf{11.03} & \textbf{53.48} & \textbf{65.70} & \textbf{0.778} & \textbf{0.797} \\
        \bottomrule
    \end{tabular}
    }

\end{table}

\begin{figure}[htbp]
    \centering
    \includegraphics[width=\linewidth]{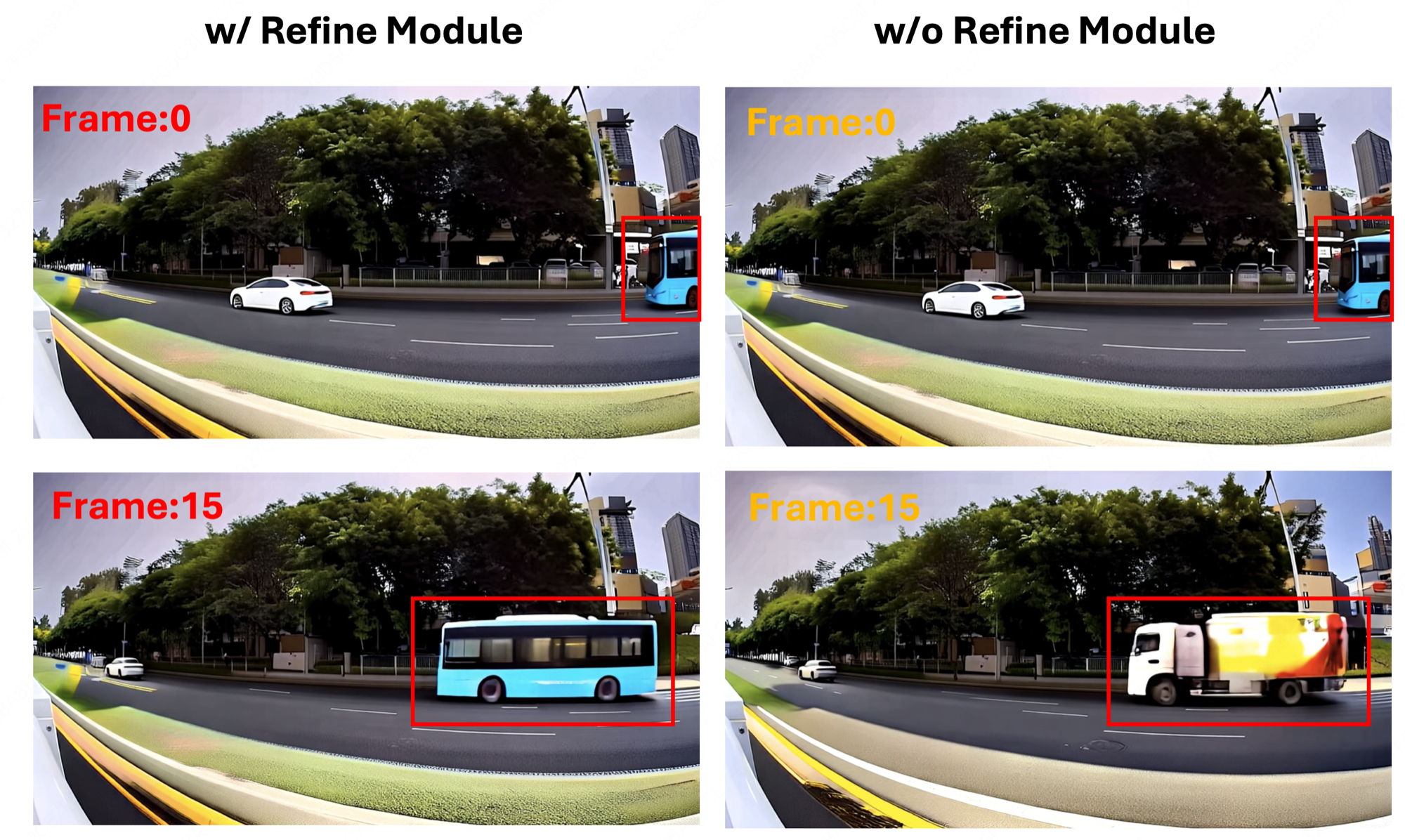}
    \caption{ The visual comparison results with (left) or without (right) using the object-level refinement module for a fast-moving vehicle generation.}  
    \label{fig:qualitative_ablation}
\end{figure}

\textbf{Object-Level Refinement Module.} To investigate the impact of our {object-level refinement module}, we compare the performance of our full system against a baseline variant without the refinement. As reported in Table \ref{tab:ablation}, the refinement module yields comprehensive improvements across all metrics, significantly boosting fine-grained control, evident in the higher \textbf{mAP} and \textbf{IoU} scores. Moreover, the substantial increase in \textbf{CLIP-T} and \textbf{CLIP-I} scores confirms that our module effectively aligns the generated objects with the specific textual and visual conditions. As illustrated in Fig. \ref{fig:qualitative_ablation}, the refinement module successfully rectifies local artifacts in the baseline outputs, such as blurred textures and inconsistent appearances across views, resulting in higher-fidelity, spatiotemporally consistent video generation.

\begin{table}[h]
    \centering
    \caption{\textbf{Ablation study of the Refinement Module.} We compare the generation quality with and without the refinement loop. The full model demonstrates superior performance in video quality, geometric precision, and semantic alignment.}
    \label{tab:ablation}
    \resizebox{\linewidth}{!}{
    \begin{tabular}{l cc cc cc}
        \toprule
        \multirow{2}{*}{\textbf{Settings}} & \multicolumn{2}{c}{\textbf{Video Quality}} & \multicolumn{2}{c}{\textbf{Layout Fidelity}} & \multicolumn{2}{c}{\textbf{Alignment}} \\
        \cmidrule(lr){2-3} \cmidrule(lr){4-5} \cmidrule(lr){6-7}
         & \textbf{FVD} ($\downarrow$) & \textbf{FID} ($\downarrow$) & \textbf{mAP} ($\uparrow$) & \textbf{IoU} ($\uparrow$) & \textbf{CLIP-T} ($\uparrow$) & \textbf{CLIP-I} ($\uparrow$) \\
        \midrule
        w/o Refinement & 108.3 & \textbf{11.01} & 48.78 & 58.82 & 0.672 & 0.691 \\
        \textbf{w/ Refinement} & \textbf{107.7} & 11.03 & \textbf{53.48} & \textbf{65.70} & \textbf{0.778} & \textbf{0.797} \\
        \bottomrule
    \end{tabular}
    }
\end{table}

\begin{figure}[htbp]
    \centering
    \includegraphics[width=\linewidth]{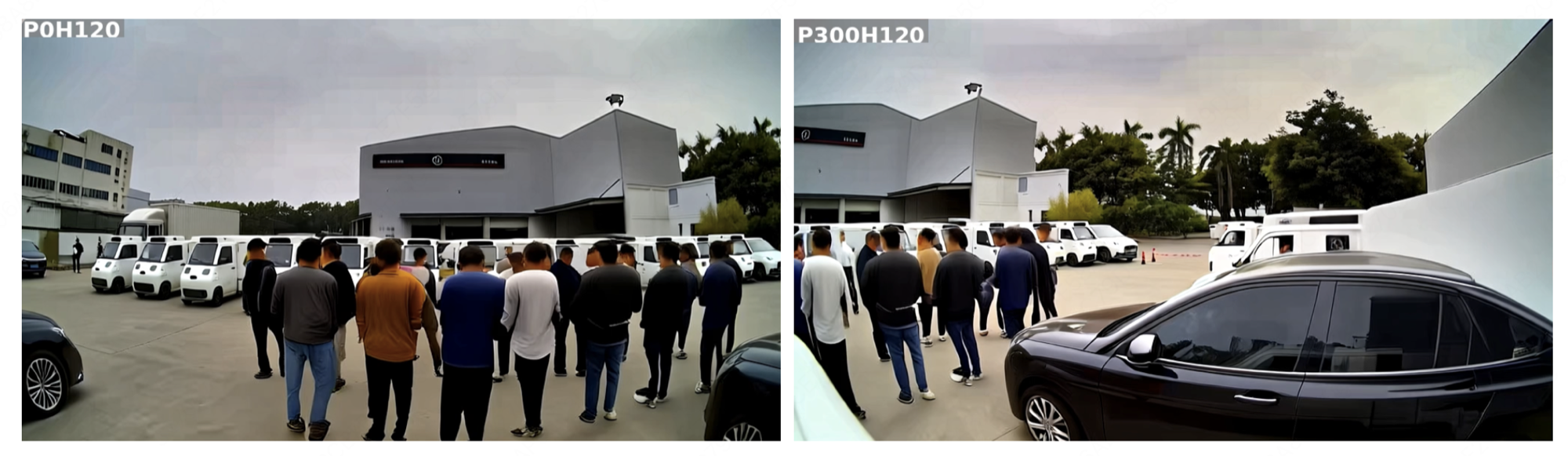}
    \caption{Inpainting capability to highly articulated or rapidly deforming objects (e.g., pedestrians in dense crowds) remains an open challenge.}  
    \label{fig:limitations}
\end{figure}

\subsection{Limitation \ssh{and Discussions}}

\clh{
Despite the compelling results, our current framework presents a few limitations that warrant future investigation. The introduction of the generation-evaluation-regeneration closed-loop mechanism inevitably introduces computational overhead during inference. Compared to single-pass, open-loop feed-forward diffusion models, the sequential auditing by the MV-VLM evaluator and the subsequent 3D-guided refinement process increase the overall rendering latency. To mitigate this, future work could explore model distillation techniques to embed the closed-loop reasoning capabilities into a more efficient single-pass generator, alongside adopting lightweight VLM backbones and parallelized evaluation strategies. Besides, as shown in Fig.\ref{fig:limitations}, while our object-level refinement seamlessly handles rigid dynamic agents (e.g., vehicles) via 3D proxy tracking, extending this precise inpainting capability to highly articulated or rapidly deforming objects (e.g., pedestrians in dense crowds) remains an open challenge. Future iterations could overcome this by integrating parametric human models (e.g., SMPL) or kinematic skeleton tracking\cite{qiao2020synthesizing} as advanced spatial proxies to accurately guide the refinement of non-rigid entities.
}

%% file: sec/4-conclusion.tex
\section{Conclusion}

This paper has presented a new driving video generation technique, called VistaGEN, which leverages visual-language reasoning cues to formulate a novel \emph{generation-evaluation-regeneration} closed-loop generation mechanism, achieving state-of-the-art fine-grained controllability and spatiotemporal consistency for long video sequence generation. We hope this work will inspire subsequent efforts exploring more powerful generative models from multimodal cues for diverse and consistent video generation across general domains.